\documentclass[runningheads]{llncs}
\usepackage[T1]{fontenc}
\usepackage{graphicx,verbatim}
\usepackage{amsmath}
\usepackage{amssymb}
\usepackage{float}
\usepackage{amsfonts}
\usepackage{algorithm}
\usepackage{algpseudocode}
\usepackage{booktabs}
\usepackage{color}
\usepackage[colorlinks=true,urlcolor=blue,linkcolor=black,citecolor=black]{hyperref}

\usepackage[misc]{ifsym}
\begin{document}
\title{
Feature-Space Guided Diffusion for Realistic Ultrasound Image Synthesis
}

\titlerunning{Feature-Space Guided Diffusion}

\author{
Marina Domínguez\inst{1}\textsuperscript{(\Letter)} \and
Nélida Mirabet-Herranz\inst{1} \and
Valery Naranjo\inst{1,2}
}

\authorrunning{M. Domínguez et al.}

\institute{
Instituto Universitario de Investigación en Tecnología Centrada en el Ser Humano,
HUMAN-tech, Universitat Politècnica de València, Valencia, Spain\\
\email{mdommar1@doctor.upv.es}
\and
Artikode Intelligence S.L., Valencia, Spain
}

\maketitle

\begin{abstract}
Conditional diffusion models can generate anatomically plausible medical ultrasound (US) images, but anatomical plausibility alone does not ensure realistic B-mode appearance. Most US pipelines adapt standard generative architectures and condition them on anatomical masks, or use guidance mechanisms that reinforce the same anatomical signal. However, B-mode US images are shaped by acquisition-dependent properties such as speckle texture, tissue contrast, and attenuation. Using a frozen US foundation model, we show that standard conditional diffusion baselines remain separated from real images in representation space. In this work, we propose \emph{Feature-Space Candidate Guidance} (FSCG), a training-free sampling strategy to reduce this gap. At sampling time, FSCG applies local k-NN feature correction and selects the best of multiple stochastic candidates according to their feature-space energy. In this way, the mask defines the anatomy, while FSCG steers samples toward the real US domain. Across three different datasets, FSCG reduces average FID64 by 56\%, FID192 by 57\%, and nearest-neighbour feature distance by 47\% over standard conditional diffusion sampling, outperforming alternative inference-time guidance baselines. The results suggest that domain-aware feature representations can reveal and reduce realism gaps in medical diffusion synthesis without retraining the generator. Our code is available at
\url{https://github.com/marinadominguez/FSCG}.


\end{abstract}

\section{Introduction}
\label{sec:intro}

Medical ultrasound (US) is a widely used clinical imaging modality due to its low cost, portability, real-time acquisition, and absence of ionizing radiation~\cite{Wang2020UltrasoundSafePortable}. However, developing robust machine-learning models for US remains challenging: image appearance varies substantially across operators, acquisition protocols, probes, anatomies, and clinical centres. Moreover, expert annotations are costly and data sharing is often limited by privacy constraints~\cite{Tupper2025USAug,Singla2022SpeckleShadows}. Realistic US synthesis could therefore support data augmentation, robustness analysis, and simulation when large curated datasets are not available~\cite{Dominguez2024DiffusionSoundPropagation,Stojanovski2023EchoFromNoise,Wu2026EchoAdapter}.

Medical image synthesis, and ultrasound image generation in particular, has increasingly benefited from diffusion models~\cite{Azad2026SystematicReview,Ibrahim2024SyntheticDataReview}. Most conditional US synthesis methods control the generated anatomy through masks, semantic maps, or related structural inputs~\cite{Stojanovski2023EchoFromNoise,Wu2026EchoAdapter}. These signals specify the spatial layout, but they do not explicitly enforce realistic B-mode appearance. Moreover, evaluating the realism of the images is challenging because common metrics, such as PSNR, SSIM and LPIPS, are sensitive to normalization, alignment, background regions, or blur~\cite{Dohmen2024FivePitfalls}, whereas distributional measures such as FID~\cite{Heusel2017FID} rely on natural-image feature spaces that do not reflect US-specific appearance characteristics~\cite{Kynkaanniemi2022RoleImageNetFID,Scholz2024GlobalConsistency}. These limitations call for evaluation and guidance in representation spaces learned from the actual ultrasound data.

\begin{figure}[!b]
\centering
\includegraphics[width=0.9\linewidth]{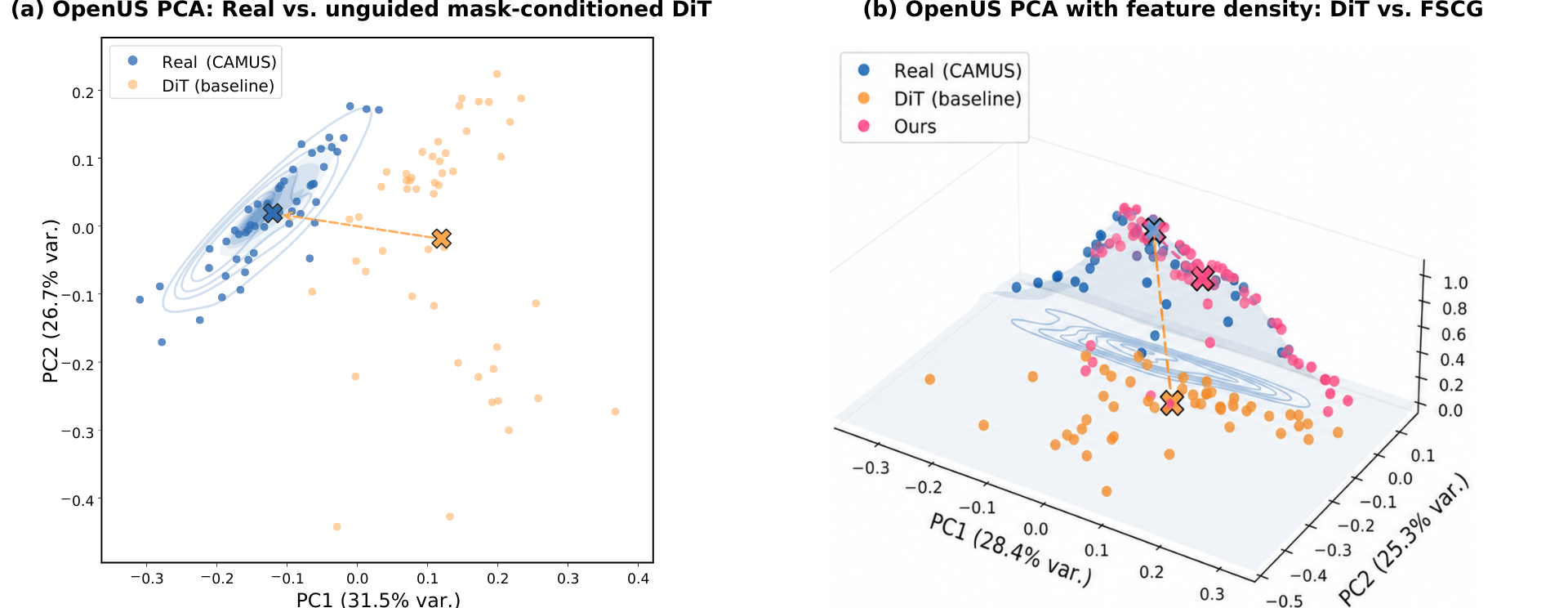}
\caption{\textbf{OpenUS feature-space analysis.}
(a) PCA projection shows a representation gap between real images and DiT samples.
(b) Applying our FSCG guidance method, samples are shifted toward the real-image distribution.
Cross markers denote centroids.}
\label{fig:final_openus_global_gap_intro}
\end{figure}

Figure~\ref{fig:final_openus_global_gap_intro} motivates our approach. We project real echocardiographic 
images~\cite{Leclerc2019CAMUS} and samples from a conditioned diffusion transformer (DiT) into the feature space of OpenUS, a foundation model for US image analysis~\cite{Zheng2025OpenUS}.
We use DiT as baseline model because it has shown excellent scalability and state-of-the-art image-generation performance on class-conditional benchmarks~\cite{Peebles2023DiT}. 
We consider a mask-conditioned DiT baseline, where a binary anatomical mask is provided as structural conditioning to specify the target cardiac geometry during generation. While this conditioning guides the reverse diffusion process toward the desired anatomy, DiT samples remain shifted from real images in feature space (Fig.~\ref{fig:final_openus_global_gap_intro}(a)).
An explanation is that standard diffusion training relies primarily on denoising objectives that do not directly encode perceptual or modality-specific realism~\cite{Ho2020DDPM,Lin2025PerceptualLoss}. This inspires the use of US-specific representations not only for evaluation but also as a source of guidance during sampling.

We propose \emph{Feature-Space Candidate Guidance} (FSCG), a training-free guidance strategy for conditional US diffusion models (Fig.~\ref{fig:final_openus_global_gap_intro}(b)). FSCG combines local OpenUS kNN feature-space correction with stochastic candidate selection during reverse diffusion, improving alignment with real US features without retraining or modifying the generator. Unlike settings where the desired constraint can be written explicitly~\cite{Christopher2024PDM}, US realism lacks a simple closed-form criterion. FSCG represents the target appearance US distribution through frozen OpenUS features extracted from real training images, and uses this representation to score and steer samples toward the region of feature space occupied by real images.

Our main contributions can be summarised as follows:
(a) We identify a lack of realism evaluation metrics and a representation gap between real US images and samples, showing that mask conditioning alone does not guarantee alignment with the real US domain.
(b) We introduce FSCG, a training-free guidance mechanism that uses a real-image OpenUS feature bank to steer reverse diffusion.
(c) We show across anatomically different US datasets that FSCG improves feature-space alignment and image-level distributional metrics over standard sampling and alternative feature-space guidance baselines, using complementary pixel, perceptual, distributional, US texture, and OpenUS feature metrics.

\section{Related Work}
\label{sec:related}

\textbf{Diffusion-based medical and US synthesis.}
Diffusion models provide a powerful framework for image generation. DDPMs established a practical probabilistic denoising formulation~\cite{Ho2020DDPM}, score-based models generalized diffusion through continuous-time stochastic dynamics~\cite{Song2021ScoreSDE}, latent diffusion reduced computational cost by operating in compressed autoencoder spaces~\cite{Rombach2022LDM}, and DiT replaced U-Net backbones with scalable transformer architectures~\cite{Peebles2023DiT}. In medical imaging, diffusion models have been explored for synthesis, reconstruction, segmentation, denoising, super-resolution, and image-to-image translation~\cite{Azad2026SystematicReview}. In ultrasound, prior work has studied image generation from semantic or anatomical conditions~\cite{Stojanovski2023EchoFromNoise}, cardiac US synthesis~\cite{Wu2026EchoAdapter}, restoration and dehazing~\cite{Stevens2024DehazingUS}, super-resolution~\cite{Liu2024DMUISR}, segmentation-oriented diffusion models~\cite{Li2026DMUSNet}, and physics-inspired generation~\cite{Dominguez2024DiffusionSoundPropagation}. These works demonstrate the adaptability of diffusion models to US imaging, but mainly through modified training objectives or conditioning mechanisms. In contrast, we keep the generator fixed and steer its reverse diffusion trajectory toward the real ultrasound distribution at inference time.

\textbf{Evaluation of synthetic medical images.}
Evaluating whether synthetic US images are realistic remains difficult. Full-reference metrics such as PSNR, SSIM, MAE, and LPIPS are useful in reconstruction settings, but they can be misleading for generated medical images because pixel similarity does not necessarily reflect clinically meaningful quality~\cite{Dohmen2024FivePitfalls}. Distributional metrics such as FID compare real and generated collections, but they rely on feature spaces learned from natural images and may overlook US-specific properties such as speckle, tissue boundaries, acoustic shadows, and probe-dependent appearance~\cite{Heusel2017FID,Kynkaanniemi2022RoleImageNetFID}. Prior work has therefore emphasized that synthetic medical images should be assessed with complementary criteria, including medical plausibility, global consistency, and domain-specific content~\cite{Scholz2024GlobalConsistency,Dominguez2024DiffusionSoundPropagation,Dohmen2024FivePitfalls}. Ultrasound foundation models such as OpenUS offer a way to measure image proximity in a feature space learned from US data~\cite{Zheng2025OpenUS}. In this work, we use this space both to quantify the gap between real and generated images and to define a guidance signal for sampling.

\textbf{Inference-time guidance for diffusion models.}
Inference-time guidance offers an alternative to retraining, steering a fixed diffusion model during sampling~\cite{Li2024SNAFusion}.  
Classifier-free guidance combines conditional and unconditional predictions~\cite{Ho2022ClassifierFreeGuidance}, while methods such as Universal Guidance and FreeDoM use external objectives or pretrained networks to guide sampling~\cite{Bansal2023UniversalGuidance,Yu2023FreeDoM}.
Constraint- and reward-based methods instead modify the trajectory with explicit constraints or rewards: PDMs project samples onto constraint sets~\cite{Christopher2024PDM}, DTM optimizes a terminal cost~\cite{Pandey2025VariationalControl}, DAS uses Sequential Monte Carlo for reward-aligned sampling~\cite{Kim2025DAS}, and TreeG selects among proposed candidates through tree-search path steering~\cite{Guo2025TreeG}. These works guide diffusion after training, but target known constraints, generic rewards, or natural-image guidance. FSCG instead builds the guidance signal from real US images' features for local correction and candidate selection in OpenUS feature space, directly targeting the US image distribution.

\section{Method}
\label{sec:method}

We propose \emph{Feature-Space Candidate Guidance} (FSCG), a training-free sampling strategy for a fixed conditional diffusion generator. The generator is first trained in the usual way, using anatomical masks as conditioning. At inference time, FSCG keeps the DiT, image decoder, and OpenUS encoder frozen, and modifies only the reverse sampling trajectory. At each guided step, FSCG first applies a local OpenUS-$k$NN gradient correction to obtain a guided latent $z_t^{\mathrm{g}}$, then samples $M$ mask-conditioned reverse candidates from $z_t^{\mathrm{g}}$ and selects the one with lowest OpenUS-$k$NN energy. Although we instantiate the feature encoder with OpenUS, FSCG can be applied to any differentiable domain-specific representation model.

\subsubsection{Conditional latent DiT generator.} Let $\mathcal{D}=\{(x_i,m_i)\}_{i=1}^{N}$ denote real US images and their corresponding anatomical masks. Images are mapped into a latent space with a pretrained image autoencoder, $z_0=E(x_0)$, while masks are resized to the latent resolution. The baseline generator is a conditional DiT trained to denoise latent variables: at each reverse step, it receives the current latent $z_t$, timestep $t$, and mask $m$, and predicts a clean latent estimate $\hat{z}_0(z_t,t,m)$.

\subsubsection{Real-image feature-bank energy.}
Let $\phi(\cdot)$ be a frozen US feature encoder. We build the feature bank, $\mathcal{B}$, from real training images only, using $\ell_2$-normalized feature vectors:
$\mathcal{B}=\{b_i\}_{i=1}^{N}$, where $b_i = \phi(x_i)/\|\phi(x_i)\|_2$.
Validation and test images are never included in the bank. Given a predicted clean image $\hat{x}_0$, we compute its normalized feature:
$ 
q = {\phi(\hat{x}_0)}/{\|\phi(\hat{x}_0)\|_2}. 
$
We then define a local $k$-nearest-neighbour energy as:
\begin{equation}
E_{\mathrm{kNN}}(\hat{x}_0)
=
\frac{1}{k}
\sum_{j \in \mathcal{I}_k(q,\mathcal{B})}
\|q-b_j\|_2^2,
\label{eq:knn_energy}
\end{equation}
where $\mathcal{I}_k(q,\mathcal{B})$ denotes the indices of the $k$ nearest real training features to $q$. Lower energy indicates that $\hat{x}_0$ is closer to the real US feature distribution. 

\begin{figure}[t]
\includegraphics[width=\textwidth, trim=0 36 0 14, clip]{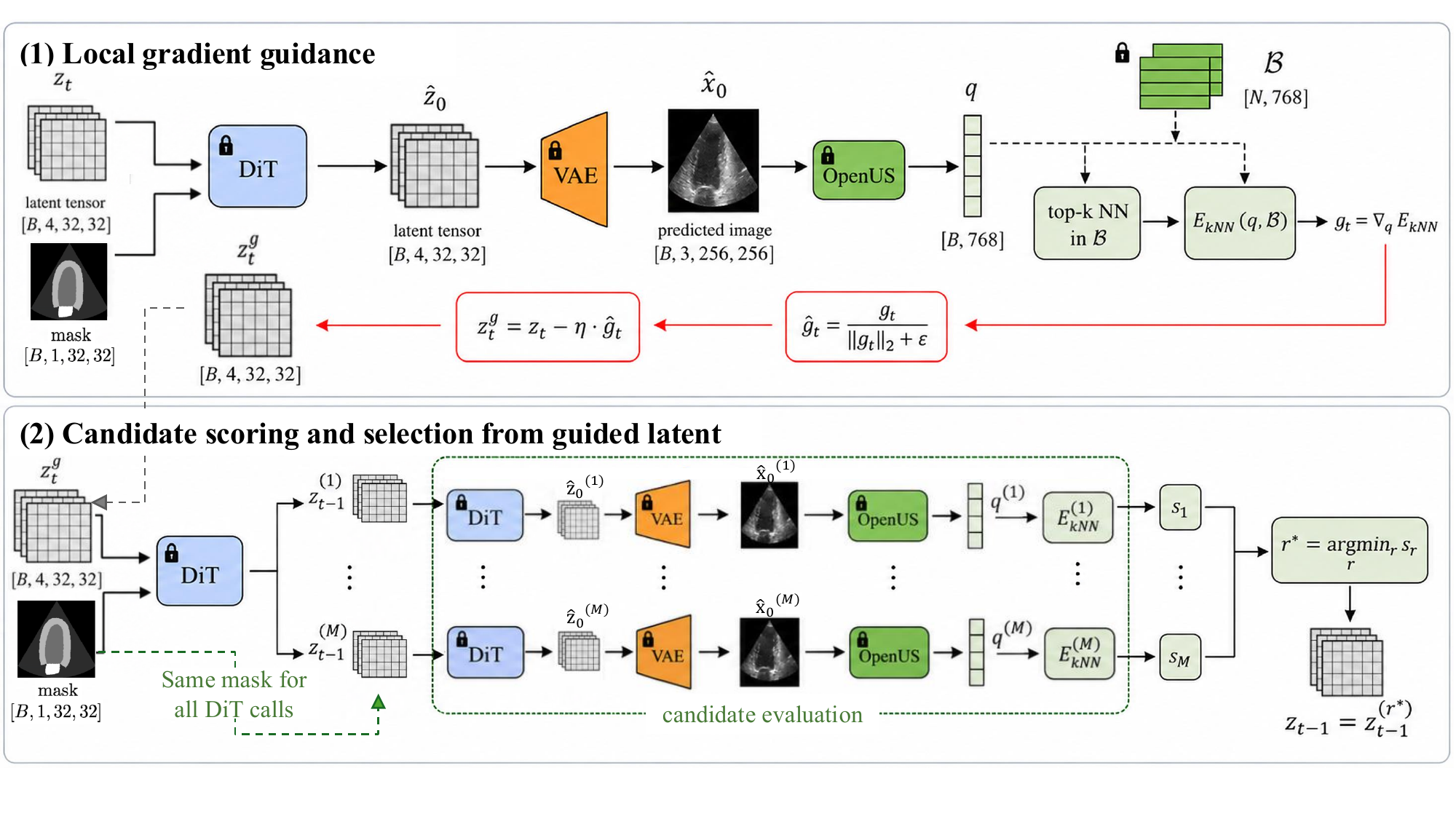}
\caption{\textbf{Feature-Space Candidate Guidance (FSCG).} (1) \emph{Local gradient guidance}: 
At selected reverse steps, the current latent $z_t$ is passed through the frozen DiT and VAE decoder to obtain a predicted clean image, which is embedded with OpenUS and compared with a real-image feature bank. The resulting local $k$NN energy provides a gradient that updates $z_t$ into $z_t^{\mathrm{g}}$. (2) \emph{Branch search from the guided latent}: 
Starting from $z_t^{\mathrm{g}}$, the frozen sampler proposes $M$ candidate reverse steps. 
The lowest-energy candidate is selected as the next latent state $z_{t-1}$.}
\label{fig:feature_space_candidate_guidance}
\end{figure}

\subsubsection{Feature-Space Candidate Guidance.}
At selected reverse timesteps, we first use the feature-bank energy for local gradient guidance. As illustrated in Fig.~\ref{fig:feature_space_candidate_guidance}~(1), the current latent $z_t$ is decoded through the predicted clean estimate, scored with Eq.~\eqref{eq:knn_energy}, and differentiated with respect to the noisy latent: 
\begin{equation}
g_t
=
\nabla_{z_t}
E_{\mathrm{kNN}}
\left(
D\left(\hat{z}_0(z_t,t,m)\right)
\right).
\end{equation}

The gradient is normalized per sample and used to obtain a guided latent, $z_t^{\mathrm{g}}
=
z_t
-
\eta
\frac{g_t}{\|g_t\|_2+\varepsilon},$
where $\eta$ controls the guidance strength and $\varepsilon$ avoids numerical instability. This step 
moves the 
trajectory toward lower feature-bank energy.

A single stochastic reverse step from $z_t^{\mathrm{g}}$ may still follow a suboptimal path. We therefore perform branch-level candidate selection (Fig.~\ref{fig:feature_space_candidate_guidance}~(2)). 
From the guided latent, the frozen DiT sampler proposes $M$ independent reverse-step candidates,

\begin{equation}
z_{t-1}^{(r)}
\overset{\mathrm{i.i.d.}}{\sim}
p_{\theta}
\left(
z_{t-1}\mid z_t^{\mathrm{g}},t,m
\right),
\qquad
r=1,\ldots,M.
\label{eq:candidate_sampling}
\end{equation}

Each candidate $z_{t-1}^{(r)}$ is first mapped to its predicted clean latent $\hat{z}_0^{(r)}$ using the frozen denoiser, decoded into image space, and scored with the same OpenUS $k$NN energy.
Let $f_{\theta}^{0}(z,t,m)$ denote the clean-latent estimate predicted by the DiT from latent $z$, timestep $t$, and mask $m$. For each candidate, we compute
\begin{equation}
\hat{z}_0^{(r)}
=
f_{\theta}^{0}
\left(
z_{t-1}^{(r)}, t-1, m
\right),
\qquad
s_r
=
E_{\mathrm{kNN}}
\left(
D\left(\hat{z}_0^{(r)}\right)
\right).
\label{eq:candidate_scoring}
\end{equation}

The next latent is selected as the lowest-energy candidate, 
\begin{equation}
r^\star
=
\arg\min_{r \in \{1,\ldots,M\}}
s_r,
\qquad
z_{t-1}
=
z_{t-1}^{(r^\star)}.
\label{eq:branch_selection}
\end{equation}

\section{Experimental Setup}
\label{sec:experimental_setup}

\subsection{Datasets}

We evaluate FSCG on three US datasets with paired masks: cardiac, breast, and ovarian (Table~\ref{tab:datasets}). All images are resized to $256\times256$, masks are used as spatial conditioning, and offline augmentation is applied only to training splits.

\begin{table}[h]
\centering
\caption{Datasets used for mask-conditioned ultrasound synthesis.}
\label{tab:datasets}
\begin{tabular}{lllc}
\toprule
Dataset & Domain / mask & Clean split & Aug. train \\
\hline
CAMUS~\cite{Leclerc2019CAMUS} & Cardiac / anatomy & 400 train, 50 val & 2000 \\
BUS\_UC~\cite{Iqbal2023BUSUCData} & Breast / lesion & 647 train, 100 test & 3235 \\
MMOTU~\cite{Zhao2022MMOTU} & Ovarian / fan+tumour & 445 train, 80 test & 2225 \\
\bottomrule
\end{tabular}
\end{table}

For CAMUS, we use two-chamber end-diastolic B-mode images and follow the standard split and offline augmentation used in previous works~\cite{Stojanovski2023EchoFromNoise}. For BUS\_UC, we remove invalid image-mask pairs and apply mild spatial, intensity, speckle, blur, and attenuation augmentations to images and masks, using NN-interpolation for masks~\cite{Tupper2025USAug,Singla2022SpeckleShadows}. For MMOTU, we use OTU-2D B-mode with a three-level mask encoding background, US fan, and tumour region. After removing invalid pairs, we apply the same augmentation protocol as for BUS\_UC.

\subsection{Experimental protocol}

\subsubsection{Model and sampling.}
All trained models are mask-conditioned DiTs~\cite{Peebles2023DiT}. Images are encoded into $4\times32\times32$ VAE latents, and masks are resized to $32\times32$ and concatenated with the noisy latent, yielding five DiT input channels. 
To ensure a fair comparison, all guidance baselines use the same OpenUS feature bank, DiT checkpoint, held-out masks, sampling schedule, and number of samples. Thus, the compared methods differ only in how they use the OpenUS feature signal during sampling. Feature banks contain non-augmented training images: 400 CAMUS, 647 BUS\_UC, and 445 MMOTU. EMA checkpoints are 34k/130k/140k for CAMUS/BUS\_UC/MMOTU. Unless stated otherwise, we use 100 DDPM steps, apply guidance in the final 10\% of steps, and set $k=5$, $M=4$, and $\eta=1.0$. On a DGX A100 system using a single GPU, 50 samples took about 40 min for FSCG and 25--27 min for most baselines.

\subsubsection{Evaluation.}
For each dataset, we generate 50 samples from held-out masks and compare them with the corresponding 50 real images. 
This number is fixed by the CAMUS validation split and kept across datasets for a matched evaluation protocol.
Since no metric captures clinical realism in synthetic US, we report complementary metrics: FID at two deeper feature resolutions~\cite{Dominguez2024DiffusionSoundPropagation}, LPIPS, US texture/frequency discrepancies, and OpenUS feature-space alignment using nearest-neighbour cosine, L2, and centroid distances. We interpret all metrics jointly with qualitative comparisons. 

\section{Results}
\label{sec:results}

Table~\ref{tab:results_main_average} summarizes the quantitative comparison of inference-time guidance strategies for mask-conditioned DiT US image synthesis. The \emph{real-reference} row reports variability between real validation images, providing a lower-bound reference scale. The remaining rows compare inference-time variants that differ only in how the feature-bank signal is used during sampling. Since samples are stochastic outputs conditioned on masks rather than deterministic reconstructions of paired real images, we emphasize distributional and representation-level metrics over pair-wise scores. FSCG obtains the lowest FID64 and FID192, and the best NN-alignment in feature space according to both cosine and Euclidean distances. Centroid guidance obtains the lowest centroid and US-texture averages, consistent with its global pull toward the centroid. However, FSCG better matches local real-image neighbourhoods while improving image-level distributional alignment, suggesting a closer fit to the real US feature distribution.

\begin{table*}[t]
\centering
\caption{
Average quantitative comparison across three US datasets. Methods are grouped by the guidance principle. US-Avg is the unweighted average of five US-specific texture discrepancies: speckle, contrast, dynamic range, edge and FFT distance.
NN-CosD. and NN-L2 denote OpenUS nearest-neighbour cosine and L2 distances, and Ctd. denotes centroid distance.
Rank is the mean rank over the reported metrics and is shown as a compact summary.
}
\label{tab:results_main_average}
\resizebox{\textwidth}{!}{%
\begin{tabular}{lcccccccc}
\toprule
Method & FID64 $\downarrow$ & FID192 $\downarrow$ & LPIPS $\downarrow$ & US-Avg $\downarrow$ & NN-CosD. $\downarrow$ & NN-L2 $\downarrow$ & Ctd. $\downarrow$ & Rank $\downarrow$ \tabularnewline 

\midrule
\textit{Real-reference}
& 0.080 & 0.396 & 0.302 & 0.033 & 0.023 & 0.197 & 0.091 & -- \tabularnewline

\midrule
\textit{Baseline}: DiT$_{\mbox{\scriptsize ICCV'23}}$~\cite{Peebles2023DiT}
& 0.524 & 2.670 & 0.330 & 0.151 & 0.112 & 0.426 & 0.751 & 8.43 \tabularnewline 

\midrule
\multicolumn{9}{l}{\textit{Gradient-based guidance}} \tabularnewline
Centroid$_{\mbox{\scriptsize ICCV'23}}$~\cite{
Yu2023FreeDoM}
& 0.333 & 1.410 & \textbf{0.313} & \textbf{0.088} & 0.037 & 0.246 & \textbf{0.374} & 2.14 \tabularnewline 

DTM$_{\mbox{\scriptsize ICML'25}}$~\cite{Pandey2025VariationalControl}
& 0.385 & 2.056 & 0.323 & 0.119 & 0.043 & 0.273 & 0.473 & 4.21 \tabularnewline 

kNN$_{\mbox{\scriptsize ICLR'23}}$~\cite{Sheynin2023KNNDiffusion}
& 0.264 & 1.277 & 0.319 & 0.100 & 0.031 & 0.229 & 0.429 & 2.29 \tabularnewline 

PDM$_{\mbox{\scriptsize NeurIPS'24}}$~\cite{Christopher2024PDM}
& 0.393 & 2.056 & 0.324 & 0.124 & 0.047 & 0.291 & 0.561 & 5.21 \tabularnewline 

\midrule
\multicolumn{9}{l}{\textit{Candidate-search guidance}} \tabularnewline
DAS-SMC$_{\mbox{\scriptsize ICLR'25}}$~\cite{Kim2025DAS}
& 0.529 & 2.702 & 0.328 & 0.149 & 0.099 & 0.402 & 0.744 & 8.00  \tabularnewline 

LiDAR$_{\mbox{\scriptsize arXiv'26}}$~\cite{Kim2026LiDAR}
& 0.388 & 2.175 & 0.322 & 0.159 & 0.094 & 0.396 & 0.723 & 6.43 \tabularnewline 

TreeG$_{\mbox{\scriptsize arXiv'25}}$~\cite{Guo2025TreeG}
& 0.416 & 2.267 & 0.325 & 0.155 & 0.084 & 0.372 & 0.656 & 6.71 \tabularnewline 

\midrule
FSCG$_{\mbox{\scriptsize Ours}}$
& \textbf{0.230} & \textbf{1.161} & 0.315 & 0.111 & \textbf{0.030} & \textbf{0.226} & 0.413 & \textbf{1.57} \tabularnewline 

\bottomrule
\end{tabular}
}
\end{table*}

\begin{figure}[t]
\centering
\includegraphics[width=0.98\linewidth]{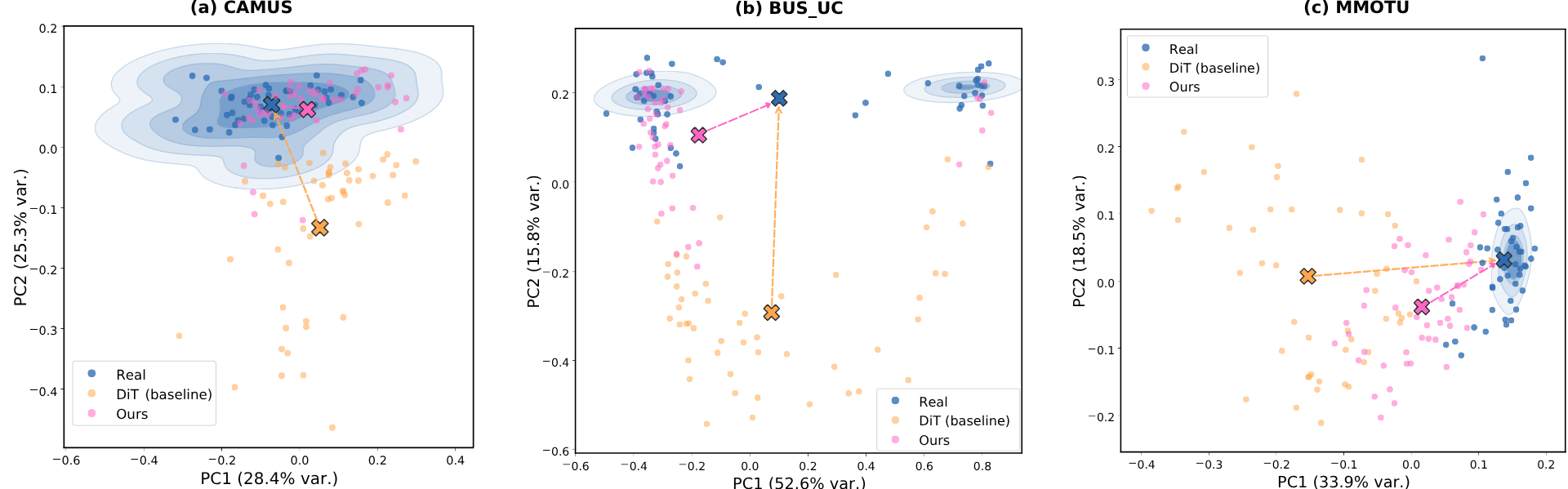}
\caption{PCA projections of real and generated samples for (a) CAMUS, (b) BUS\_UC and (c) MMOTU. Blue density contours indicate the real feature distribution, crosses denote distribution centroids, and dashed lines show centroid displacement.
}
\label{fig:final_results_openus_pca}
\end{figure}

Figure~\ref{fig:final_results_openus_pca} provides a qualitative view of the feature-space behaviour behind the quantitative metrics. Across the three datasets, unguided DiT samples remain visibly displaced from the real distribution, whereas our method reduces this gap and places more samples within or near high-density real-feature regions. Importantly, the samples are not collapsed onto a single centroid, which supports the intended effect of the local feature-bank guidance: improving proximity to real 
appearance while preserving sample variability around the conditioning mask.

\begin{figure}[t]
\centering
\includegraphics[width=\linewidth]{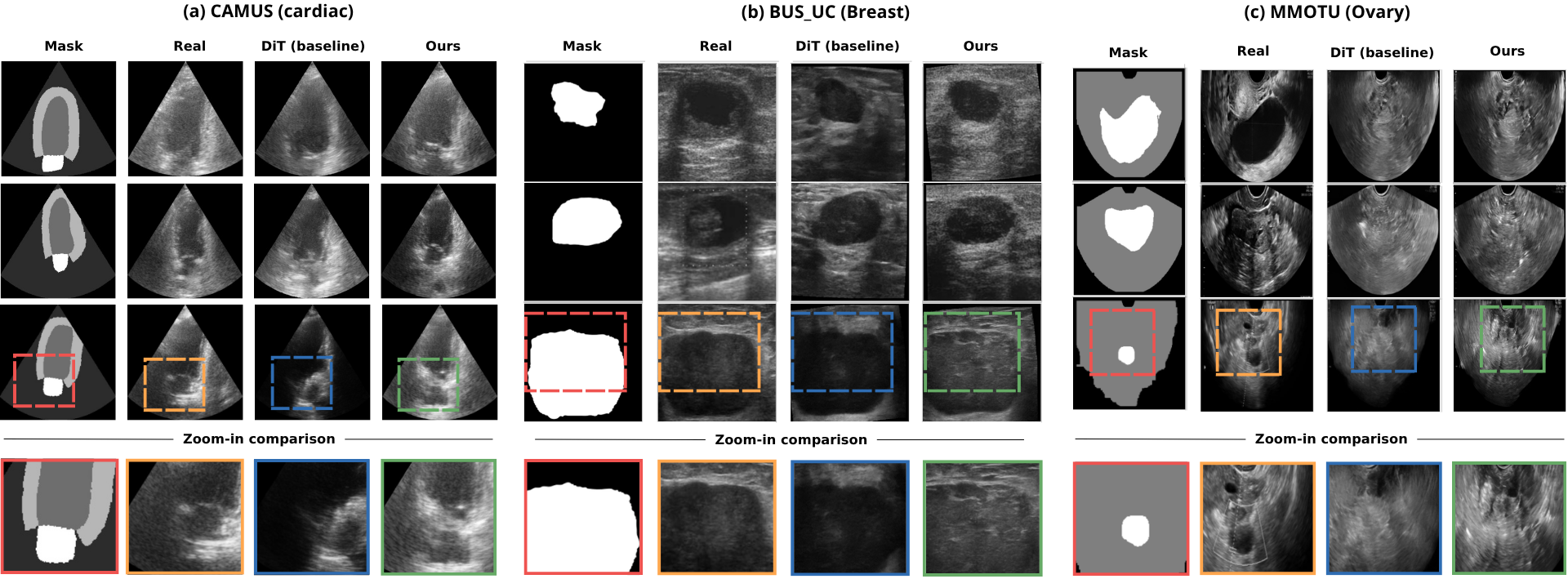}
\caption{
Qualitative comparison under matched conditioning masks across three US domains. Each panel shows the mask, real image, DiT baseline and our method. Dashed boxes indicate the regions enlarged below. Our method preserves the mask anatomy while improving local US appearance: speckle, cavity/lesion borders, and local contrast.
}
\label{fig:final_results_qualitative}
\end{figure}

Figure~\ref{fig:final_results_qualitative} confirms the same trend at the image level. The unguided DiT baseline generally follows the conditioning mask, but often produces over-smoothed regions, weak local contrast, or less convincing tissue texture. Our method keeps the anatomical layout fixed while improving local echogenic detail, speckle-like texture, and boundary definition across the different US examples. Together with the PCA analysis, this suggests a clear improvement in US-domain appearance.

\section{Conclusion}
\label{sec:conclusion}

Reliable synthetic ultrasound generation requires not only visually plausible samples, but also alignment with the structure and texture of real acquisitions. In this work, we showed that conditional diffusion models can remain misaligned with real US acquisitions when evaluated in a domain-specific feature space, despite producing plausible images. To reduce this gap, we introduced FSCG, a training-free sampling strategy that uses a frozen US foundation model to steer generation toward the real US domain. Across datasets, FSCG improved distributional alignment, feature-space proximity, and qualitative realism over unguided sampling and alternative guidance baselines. 
Future work will extend this idea to downstream tasks and domain-specific realism assessment metrics.

\begin{credits}
\subsubsection{\ackname}
IMPLANTEU is funded by the European Union's Horizon Europe research and innovation programme under the Marie Sk{\l}odowska-Curie Actions Doctoral Network (MSCA-DN) [Grant Agreement No. 101169308]. Views and opinions expressed are however those of the author(s) only and do not necessarily reflect those of the European Union or the European Commission. Neither the European Union nor the granting authority can be held responsible for them.

\subsubsection{\discintname}
The authors have no competing interests to declare.
\end{credits}

%
%
%
\bibliographystyle{splncs04}
\bibliography{references}
%

\end{document}